\documentclass[journal]{IEEEtran}

\usepackage{amsmath}					
\usepackage{amsfonts}					
\usepackage{amssymb} 					
\usepackage{cite}						
\usepackage{url}						
\usepackage{caption}					
\usepackage{graphicx}					
\graphicspath{{./}}			
\usepackage{multirow, makecell}			
\usepackage{booktabs}					
\usepackage{tabularx}					
\usepackage{threeparttable}				
\usepackage{hyperref}					

\begin{document}
\title{3D IoU-Net: IoU Guided 3D Object Detector for Point Clouds}
\author{Jiale Li, Shujie Luo, Ziqi Zhu, Hang Dai, Andrey S. Krylov, Yong Ding, and Ling Shao
\thanks{This work was supported by the National Key Research and Development Program of China (2018YFE0183900). (\textit{Corresponding author: Hang Dai, Yong Ding.})\par
J. Li, S. Luo, Z. Zhu and Y. Ding are with College of Information Science and Electronic Engineering, Zhejiang University, Hangzhou, China (email: jialeli@zju.edu.cn; luoshujie@zju.edu.cn; misteriers@foxmail.com; dingy@vlsi.zju.edu.cn).\par
A. S. Krylov is with Laboratory of Mathematical Methods of Image Processing, Lomonosov Moscow State University, Moscow, Russia (e-mail: kryl@cs.msu.ru).\par
Hang Dai and Ling Shao are with MBZUAI and IIAI in Abu Dhabi, United Arab Emirates (e-mail: hang.dai@mbzuai.ac.ae, ling.shao@ieee.org).}
}


\markboth{Journal of \LaTeX\ Class Files, Preprint,~Vol.~XX, No.~XX, APRIL~20XX}
{: }

\maketitle

\begin{abstract}
	LiDAR point clouds is essential but challenging for 3D object detection. Most existing methods focus on classification and bounding box regression in the way of designing effective network to learn expressive feature from this irregular data. However, there exists another bottleneck, which is to have a good detection confidence score to correctly pick out the best matching one from several bounding boxes corresponding to the same ground truth bounding box. In this paper, we propose a novel 3D object detection framework 3D IoU-Net aimed to perceive an accurate detection confidence score, that is, the 3D Intersection-over-Union (IoU) between the predicted bounding box and the corresponding ground truth bounding box. Specifically, the proposed 3D IoU-Net generates proposals and point-wise features from raw point cloud. Give a 3D proposal, the IoU sensitive feature is pooled by our Attentive Corner Aggregation (ACA) module and Corner Geometry Encoding (CGE) module for extra 3D IoU prediction. By considering the visible parts varies from the point cloud gathering angle, the ACA module aggregate the point-wise feature from each corner’s perspective with different attention to generate a more unified feature. Besides, the novel CGE module encodes the geometric information of bounding box itself, which is always neglected in other methods. In addition, the IoU alignment operation further boosts the 3D IoU prediction. Thanks to accurate 3D IoU prediction value as detection confidence, the better localized bounding boxes are reasonably prevented from being suppressed. Experiments on KITTI car detection benchmark show that 3D IoU-Net achieves state-of-the-art performance.
\end{abstract}


\section{Introduction}
Recent years have witnessed remarkable progress in image-based 2D vision tasks like object detection \cite{FastRCNN, FasterRCNN}, arguably benefiting from the advancement of deep convolutional networks. However, at the same time, with the development of autonomous driving \cite{KITTIDataset}, 3D object detection has also been receiving increasing attention from both industry and academia. Different from 2D object detection \cite{2D_OD_TMM, 2D_OD_TIP1, 2D_OD_TIP2}, which only locates the object on the image plane, 3D object detection outputs the 3D position coordinates, 3D size, and orientation of the object in the form of a 3D bounding box.

In autonomous driving, depth data over long-range distances is necessary for the 3D object detection task. Consequently, LiDAR sensors are widely used to capture the 3D scenes and generate point clouds by scanning surrounding objects. Point clouds provide precise depth data through sparse and unordered points instead of dense depth maps \cite{Depthimage_TMM,Depthimage_TIP1,Depthimage_TIP2}. Most point cloud based 3D object detection methods focus on performing classification and regression of proposal bounding boxes from this unstructured data. Some methods  \cite{MV3D,AVOD,PIXOR,HDNET} project the point cloud as pseudo images, such as the bird’s eye view (BEV) image or the front view (FV) image, for standard 2D convolutional processing. Other methods \cite{VoxelNet,Second,PointPillars} divide the point cloud into small even voxels and apply a 3D convolution to handle the voxelized 3D space, with good results. Besides, there are yet other methods \cite{F-PointNets,PointRCNN,STD} that achieve better 3D object detection performance by adapting PointNet \cite{PointNet} or PointNet++ \cite{PointNet++} to learn point-wise features directly from the raw point cloud.

However, almost all of these 3D object detection methods based on different point cloud representations require Non-Maximum Suppression (NMS), which is a critical post-processing procedure to suppress redundant bounding boxes based on the order of detection confidence. Further, both the 3D detector \cite{STD} and the 2D detectors \cite{IoUNet,IoUUniform} have found that predicted Intersection-over-Union (IoU) is more suitable than classification confidence as the detection confidence for NMS. The IoU is a metric measuring the location accuracy of the bounding box, which is the ratio of intersection and union between bounding box A and bounding box B. Even if containing the same object, two bounding boxes with different sizes and orientations have different IoUs. We conduct an experiment in which we feed the ground truth IoU to NMS for ideal duplicate removal. As shown in Table \ref{theoreticalupperbound}, the 3D mAP performance of PointRCNN \cite{PointRCNN} on the KITTI \cite{KITTIDataset} \textit{val} split set increases from 81.63\% to 90.91\%, suggesting that the detection accuracy can be suppressed by irrelevant detection confidence. The guided IoU is calculated between the detected bounding box and the corresponding ground truth bounding box. Therefore, in order to achieve high-performance 3D object detection, we propose an IoU guided network that aims at perceiving a more accurate IoU as the detection confidence.

\begin{figure}[!t]
	\centering
	\includegraphics[width=\linewidth]{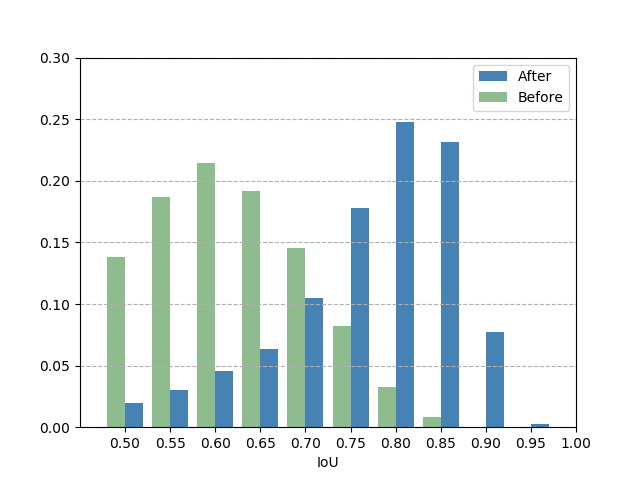}%
	\caption{IoU distribution histograms of proposal bounding boxes with their matched ground truth bounding boxes before and after refinement regression}
	\label{Histogram}
\end{figure}

There are still two problems in terms of IoU prediction. Firstly, all of the anchor-based 3D and 2D detectors \cite{STD,IoUNet,IoUUniform} directly predict the IoU by merely integrating a parallel IoU prediction head into the original network, without employing any specifically designed modules to learn the IoU sensitive features for IoU prediction. The second problem is the IoU prediction value assignment mismatching in the IoU prediction branch, as pointed out in \cite{IoUUniform}. During training, the IoU predictor learns the IoU between the proposal bounding box and the ground truth bounding box through the local feature at the proposal position. However, when it comes to the inference stage, the predicted IoU is assigned to the refined bounding box, which has been moved to a new position by the regression branch after the refinement. In Fig. \ref{Histogram}, we show the IoU distribution histogram of proposal bounding boxes before refinement regression and refined bounding boxes with their corresponding ground truth, which are calculated from the results of PointRCNN on the KITTI \textit{val} split set with IoU above 0.5. Obviously, the IoU distribution after refinement is skewed towards one side, where the IoU is close to 1, indicating that the proposal position offsets do exist and such assignment mismatching cannot be neglected.

\begin{table}[!t] 
	\caption{Theoretical detection performance upper bound of \cite{PointRCNN}.}
	\centering
	\label{theoreticalupperbound}
	\begin{tabular}{ccccc}
		\toprule	
		\multirow{2}{*}{Confidence} & \multicolumn{3}{c}{3D Object Detection AP}                         & \multirow{2}{*}{mAP} \\
		& Easy                 & Moderate             & Hard                 &                      \\
		\midrule							
		Classification score        & 88.88                & 78.63                & 77.38                & 81.63                \\
		Ground truth IoU            & 90.91                & 90.91                & 90.91                & 90.91                \\
		\bottomrule
	\end{tabular}
\end{table}

To address the first problem, we propose two modules, named the Attentive Corner Aggregation (ACA) module and Corners Geometry Encoding (CGE) module, to conduct the IoU sensitive feature learning for each proposal bounding box in our two-stage 3D IoU-Net. Given that the point clouds in autonomous driving are gathered from a fixed perspective, the objects near the LiDAR sensor are only partially covered. However, the IoU is extremely sensitive to the relationship between the visible part and full view of an object in a point cloud scene, which varies depending on the perspective. To obtain a perspective-invariant prediction head, we propose an ACA module by aggregating a local point cloud feature from each perspective of eight corners and adaptively weighting the contribution of each perspective with a perspective-channel attention mechanism. Besides, geometry information is also embedded by the proposed CGE module, which to the best of our knowledge, is the first time it has been introduced in proposal feature learning. These two feature parts are then adaptively fused by a multi-layer perceptron (MLP) network as our IoU sensitive feature. For the second problem, the IoU alignment operation is introduced to resolve the IoU assignment mismatching in the IoU prediction, without requiring any extra training. That is, the bounding box after refinement is sent back into the IoU prediction branch for a second inference, and the output is assigned to the refined bounding box as the final IoU prediction value. Experimental results demonstrate that our 3D IoU-Net achieves a better IoU perception by solving the above two problems.

The main contributions can be summarized as four-fold:
\begin{itemize}
	\item We propose a two-stage 3D object detection method, named 3D IoU-Net, with an ACA module, CGE module and IoU alignment operation. The proposed method achieves state-of-the-art performance on the highly competitive KITTI 3D object detection benchmark \cite{KITTIleaderboard}.
	\item We propose a novel feature pooling module, named the ACA module, to obtain a perspective-invariant prediction head. The existing IoU prediction methods suffer from the perspective variance. The proposed ACA module aggregates a local point cloud feature from each perspective of eight corners in the bounding box and adaptively weights the contribution of each perspective with a novel perspective-wise and channel-wise attention mechanism. 
	\item Unlike the IoU prediction in most existing methods, we use a CGE module to encode the geometry information of the bounding box, which is crucial to IoU perception. The geometry information embedding enhances the IoU sensitive feature learning. 
	\item IoU assignment mismatching is a neglected problem that constrains the performance of IoU prediction. We employ an IoU alignment operation to resolve the IoU assignment mismatching for IoU-guided NMS.
\end{itemize}

The rest of the paper is organized as follows. Section II introduces some related works on point cloud based 3D object detection. The proposed 3D IoU-Net model is detailed in Section III. Section IV shows experimental results and ablation studies. Finally, conclusions are presented in Section V.

\section{Related Works}
According to the different representations of the point cloud, 3D object detection methods based on point clouds can be roughly divided into three branches: multi-view image, voxel, and point based methods.

\subsection{3D Object Detection Based on multi-view images}
To deal with the irregularities of a point cloud, some methods \cite{MV3D,AVOD,PIXOR,HDNET,Contifuse,MMF} project it into muli-view images like BEV image and the FV image, encoded with height, density, and other statistics as the pseudo image channels. Chen \textit{et al.} \cite{MV3D} firstly turned the point cloud into a BEV image for proposal generation network (RPN) training and then pooled proposal features from the camera image, as well as both the FV image and BEV image of the point cloud for proposal refinement. Ku \textit{et al.} \cite{AVOD} improved \cite{MV3D} to achieve a higher recall proposal generation on small object instances by merging camera image and BEV image features for RPN. For higher computational efficiency, Yang \textit{et al.} \cite{PIXOR} proposed a single-stage network only based on the BEV image of the point cloud. Later, \cite{Contifuse,MMF} developed better and finer fusion strategies for multiple views. Although the methods based on the multi-view image representation mentioned above can use a 2D CNN directly, they are always limited by the information loss introduced by being projected into a certain fixed-resolution 2D grid and the handcrafted input feature channels.

\subsection{3D Object Detection Based on Voxels}
In \cite{VoxelNet}, the point cloud was evenly divided into small 3D voxels in \textit{XYZ} directions for 3D convolutional processing, and a VEF layer was proposed to adaptively learn the most expressive features of internal points in each voxel. To overcome the computational burden of the 3D CNN, Yan \textit{et al.} \cite{Second} introduced a sparse convolution \cite{sparseconv} to voxel feature learning, and Zhou \textit{et al.} \cite{DynamicVoxel} proposed a dynamic voxelization for point cloud division. Instead of dividing voxels in three directions, Lang \textit{et al.} \cite{PointPillars} generated several pillars, perpendicular to the BEV plane by ignoring the \textit{X} direction. Based on the voxel representation, the image features can be trimmed to complement the point cloud feature voxel-wise \cite{mvx}, the attention mechanism can be considered to strengthen the robustness of voxel feature learning \cite{tanet_arxiv} and so on \cite{PartA2_TPAMI,voxelfpn}. If the voxel size is too small, the many empty voxels introduce computational redundancy. However, the surface structure of the object can be destroyed if the voxel size is too large.

\subsection{3D Object Detection Based on Points}
With the introduction of PointNet \cite{PointNet} and PointNet++ \cite{PointNet++}, there is a new way of handling irregular and unordered point sets, which involves directly learning the point-wise feature from the raw point cloud without any representational transform. Qi \textit{et al.} \cite{F-PointNets} used frustum proposals from 2D detection on the corresponding camera image to narrow the search scope in a point cloud, and directly regressed predictions based on interior points. In such a cascaded framework \cite{F-PointNets,FConvNet}, the performance of 3D object detection is severely limited by the result of 2D detection. Unlike generating proposals from the image \cite{F-PointNets,FConvNet}, PointRCNN \cite{PointRCNN} generated 3D proposals from the whole point cloud with high recall performance. It directly adapted the PointNet++ to perform 3D semantic segmentation of a point cloud and generated 3D proposals based on each foreground point simultaneously. These 3D proposals were then used to pool the point-wise semantic features for further refinement. In order to improve the orientation coverage of the cubic anchor, a novel spherical anchor for point cloud space was proposed in \cite{STD}. Besides, several other methods \cite{PIRCNN,PointRGCN} also achieved better 3D object detection performance by adapting PointNet++ to point cloud processing.

Both multi-view image and voxel based methods depend on the empirical quantization resolution for point cloud representational transforms. Point-wise learning can make better use of each point in the raw point cloud, so we also conduct 3D object detection based on points.

\begin{figure*}[!th]
	\centering
	\includegraphics[width=\linewidth]{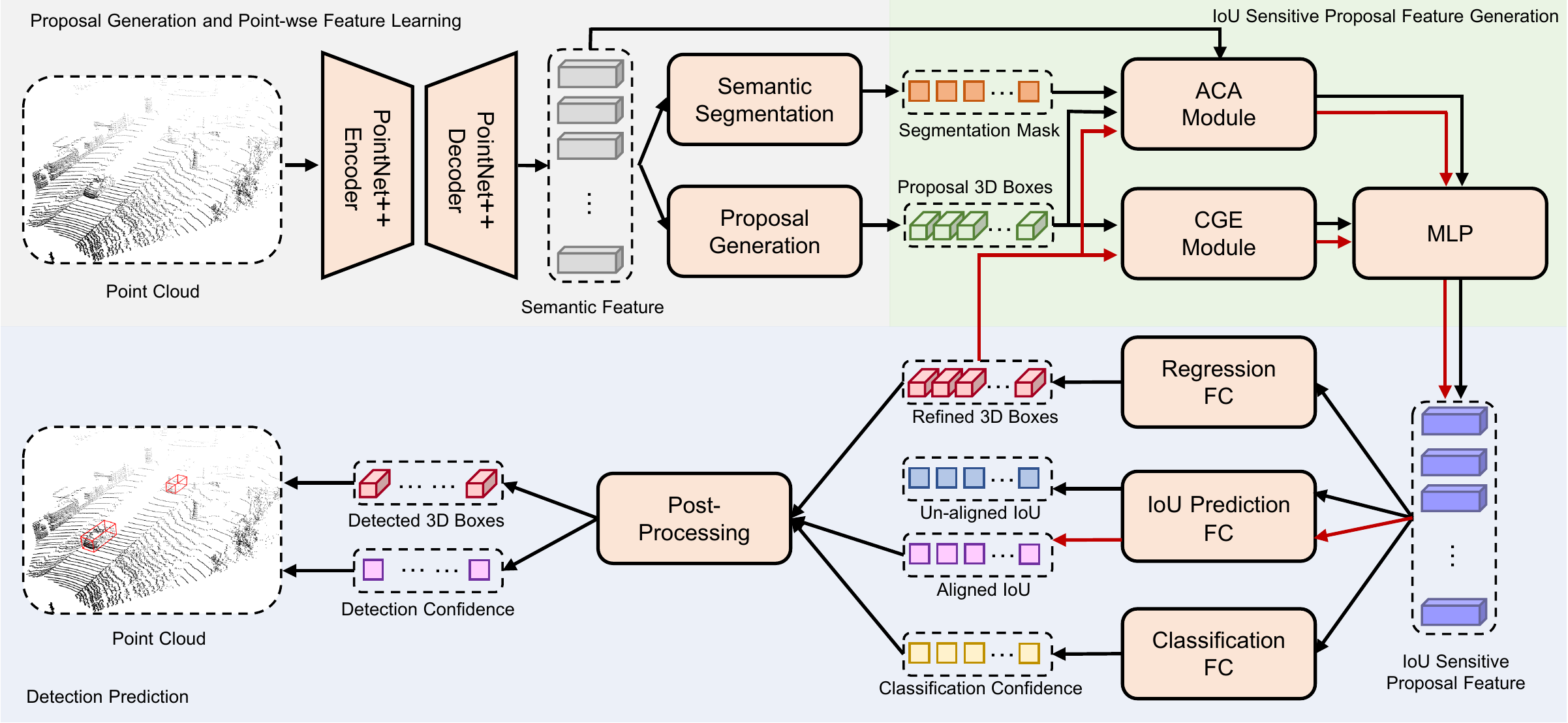}%
	\caption{Framework of proposed 3D object detection network, 3D IoU-Net. Black arrows represent the first inference, while red arrows represent the second inference brought by the IoU alignment operation. The figure is best viewed in color.}
	\label{Framework}
\end{figure*}

\begin{figure}[t]
	\centering
	\includegraphics[width=\linewidth]{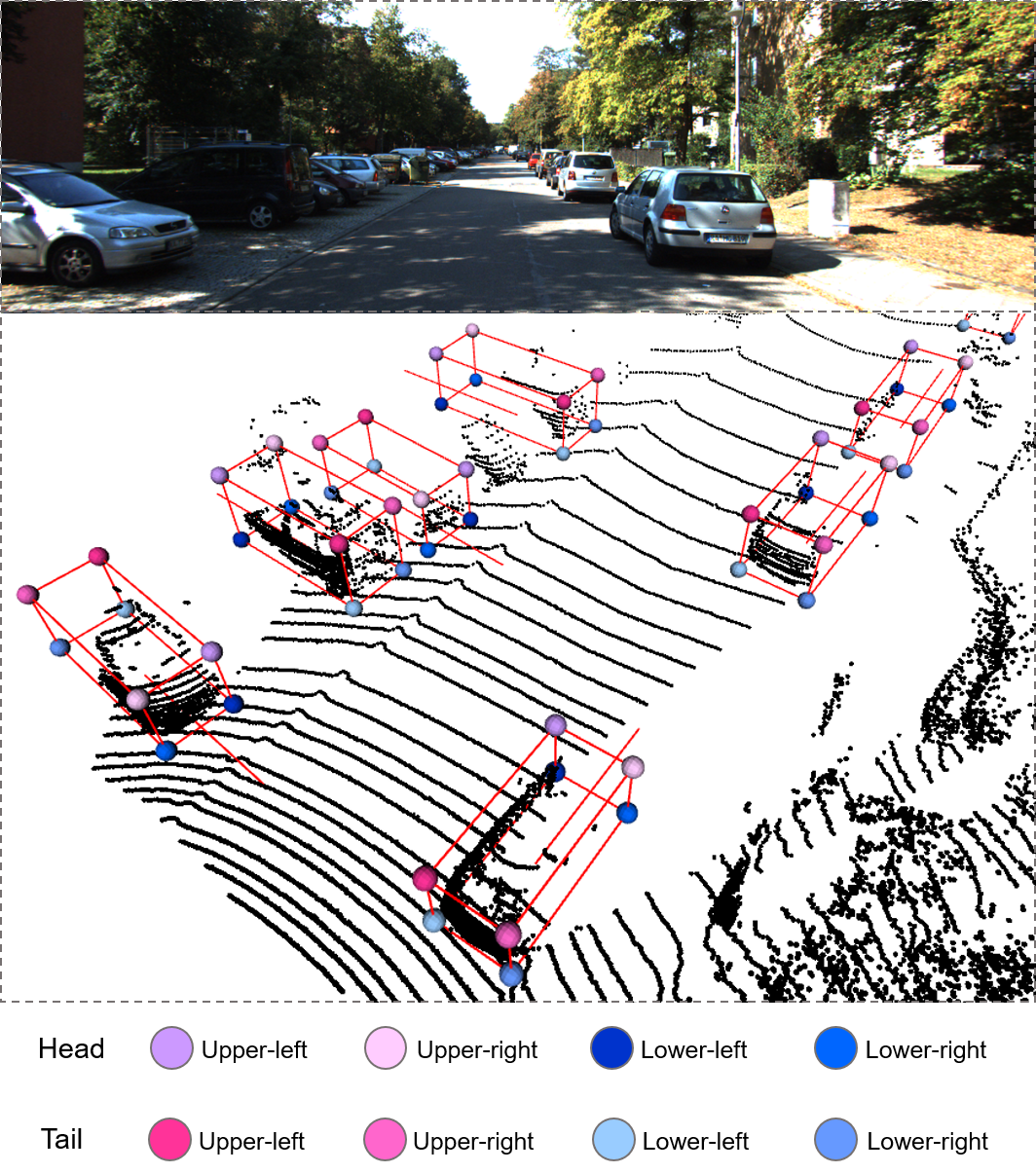}%
	\caption{Illustration of the variability in the relationship between the visible part and full view of objects. The top image is the camera image for visualization, and the bottom image is the corresponding point cloud. The points in the point cloud and the ground truth bounding boxes are colored black and red, respectively. The lines extending from the bottom of bounding boxes point to objects’ heading direction. For better observation, we zoomed in and rotated the point cloud appropriately.}
	\label{visibelpart}
\end{figure}

\begin{figure}[t]
	\centering
	\includegraphics[width=\linewidth]{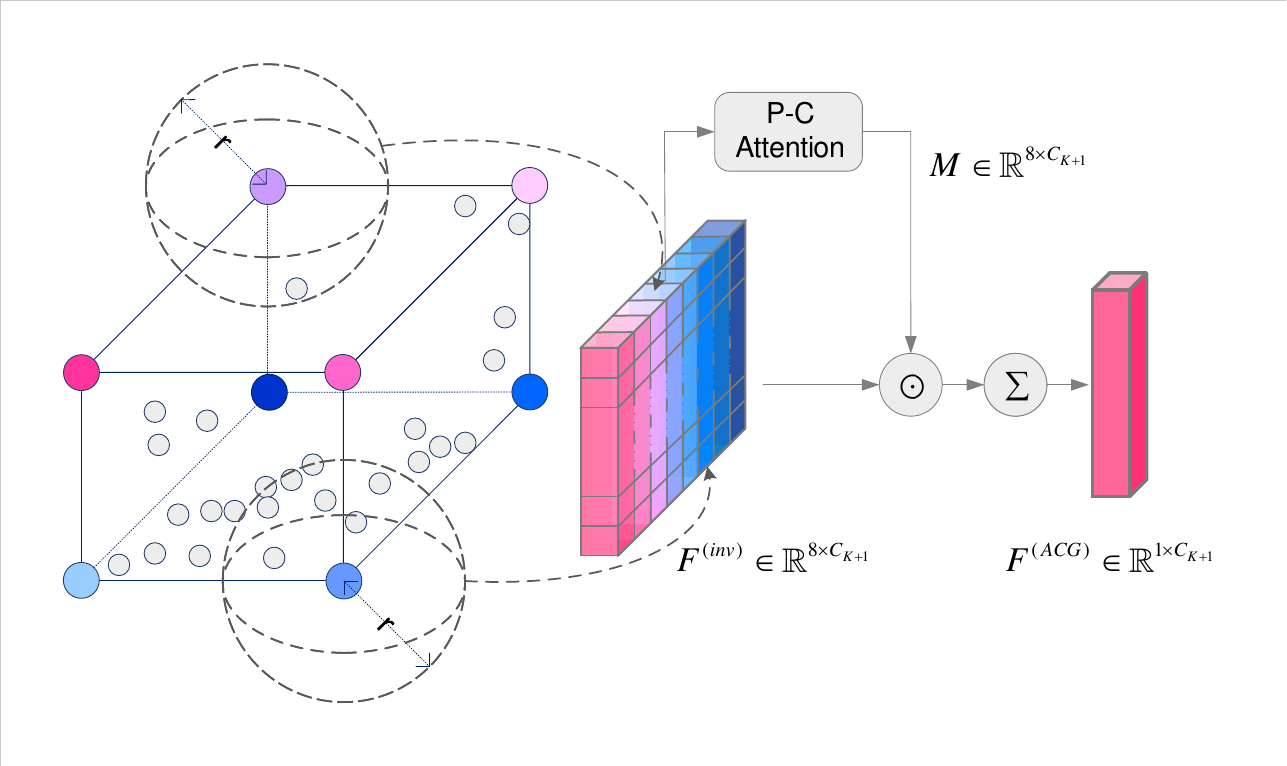}%
	\caption{Illustration of the Attentive Corner Aggregation module. The eight corners are colored differently. The $\odot$ and the $\sum$ denote the element-wise multiplication and the sum over eight corners, respectively.}
	\label{ACAmodule}
\end{figure}

\begin{figure}[t]
	\centering
	\includegraphics[width=\linewidth]{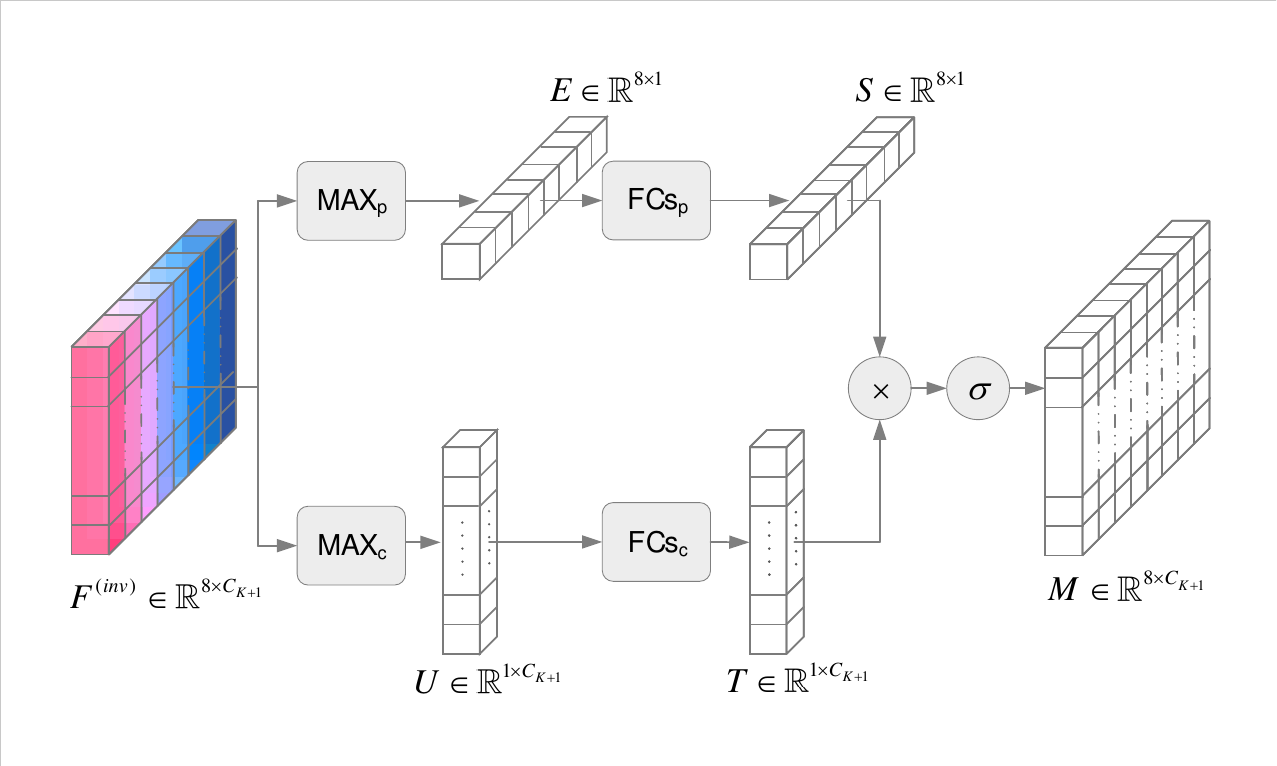}%
	\caption{Illustration of perspective-channel attention. Subscript $P$ and subscript $C$ represent perspective-wise and channel-wise, respectively. The $\times$ and the $\sigma$ denote the matrix multiplication and the sigmoid function, respectively.}
	\label{PCattention}
\end{figure}

\section{Proposed Method}
This section introduces the proposed 3D IoU-Net, a two-stage 3D detection framework. As illustrated in Fig. \ref{Framework}, 3D IoU-Net consists of a PointNet++ \cite{PointNet++} based encoder-decoder as the backbone for point-wise semantic feature learning and proposal generation. For each generated proposal, two modules, named the Attentive Corner Aggregation module and Corner Geometry Encoding module, are proposed to learn the proposal-wise IoU sensitive features. A box detection prediction network with an IoU alignment operation is applied for final box prediction.

\subsection{Proposal Generation and Point-wise Feature Learning}
We leverage the region proposal network (RPN) for 3D object proposal generation. We employ a PointNet++ based encoder-decoder as the backbone in RPN to perform point-wise segmentation from the raw point cloud and extrapolate a proposal bounding box for each foreground point. As such, the point-wise semantic features can be learned at the same time.

Given a 3D proposal, the bounding box is first expanded to include extra context surrounding the proposal and the points inside the expanded bounding box are cropped as the local point cloud. Then, the local point cloud is transformed into a canonical reference frame \cite{PointRCNN} by being subtracted from the proposal’s center and rotated to match the proposal’s orientation. For each local point $p_i$, the canonical coordinates $x_{i} ^{(pc)}$, the depth $d_i$ and the segmentation mask $s_i$ are concatenated together and embedded into a local feature $ f_i^{(loc)}$ of higher dimension by the fully connected layers $FCs_1 \left( \cdot \right)$. Then, the fully connected layers $FCs_2 \left( \cdot \right)$ are applied to merge the embedded feature with the corresponding semantic feature $f_i^{(sem)}$ for RPN as the point-wise point cloud feature $f_i^{(pc)}$, as follows:
\begin{align}
f_i^{(loc)} &= FC{s_1}(CAT(x_i^{(pc)},{d_i},{s_i})), \\ 
f_i^{(pc)} &= FC{s_2}(CAT(f_i^{(loc)},f_i^{(sem)})),
\end{align}
where $CAT \left( \cdot \right)$ represents the concatenation. Thus, the $N$ points local point cloud can be defined as
\begin{align}
{X^{(pc)}} &= [x_1^{(pc)},...,x_N^{(pc)}] \in {{\mathbb{R}}^{N \times 3}},\\
{F^{(pc)}} &= [f_1^{(pc)},...,f_N^{(pc)}] \in {{\mathbb{R}}^{N \times {C_0}}},
\end{align}
where $X^{(pc)}$ and $F^{(pc)}$ indicate the 3-dimensional coordinates and the $C_0$-dimensional point-wise point cloud features, respectively.

\subsection{IoU Sensitive Proposal Feature Generation}
The ACA and CGE modules are proposed to learn the IoU sensitive features from the local point cloud. These are shown as the green part in Fig. \ref{Framework}.
\subsubsection{Attentive corner aggregation}
The motivation behinds the ACA module is to eliminate the variability in the relationship between the visible part and full view of objects. As shown in Fig. \ref{visibelpart}, when objects are located in different positions or facing different directions, the visible part in the point cloud scene is varies. Such variability is not conducive to learning their common characteristics and perceiving the IoU. The ACA module attempts to generate a local feature from the eight-corner perspectives to perceive the full view of an object.

Details of the ACA module are illustrated in Fig. \ref{ACAmodule}. For each local point cloud, the input of the ACA module is the $X^{(pc)}$ and $F^{(pc)}$ of the local point cloud, and the ACA module outputs the proposal-wise features in three steps. Firstly, $K$ stacked Set Abstraction (SA) operations \cite{PointNet++} are employed to conduct the local point cloud feature learning, assuming that $l_k$ represents the $k$-$th$ level of the SA operation. In the $k$-$th$ SA operation, the $n_k$ intermediate points $x^{(l_k)}$ are sampled from $n_{k-1}$ intermediate points $x^{(l_{k-1})}$ by the Furthest-Point-Sampling (FPS) algorithm. Next, for each intermediate point $x_i^{(l_k)}$, $m_k$ neighboring points from $x^{(l_{k-1})}$ within radius $r_k$ are grouped and normalized by $x_i^{(l_k)}$ in order to be concatenated with the corresponding point cloud feature $f_j^{(l_{k-1})}$ as
\begin{align}
{f'}_{ij}^{({l_k})} &= CAT(x_j^{({l_{k - 1}})} - x_i^{({l_k})},f_j^{({l_{k - 1}})})\left| {\left\| {x_j^{({l_{k - 1}})} - x_i^{({l_k})}} \right\|} \right. < {r_k},\\
{f'}_i^{({l_k})} &= [{f'}_{i1}^{({l_k})},...,{f'}_{ij}^{({l_k})},...,{f'}_{i{m_k}}^{({l_k})}],
\end{align}
where $j=1, 2, ..., m_k$ and $i=1, 2, ..., n_k$. For the first level SA, $x^{(l_0)}$ and $f^{(l_0)}$ are the input $x^{(pc)}$ and $f^{(pc)}$, respectively. A PointNet \cite{PointNet} is applied to each group for higher dimensional feature learning from $C_{k-1}$ to $C_k$ as
\begin{align}
f_i^{({l_k})} = MAX(G({f'}_i^{({l_k})})),
\end{align}
where $G(\cdot)$ is an MLP network for embedding the concatenated feature ${f'}^{(l_k)}$, and $MAX(\cdot)$ is the max pooling for most expressive features along the point-axis. In this manner, each local point cloud feature is encoded into $F^((l_K)) \in \mathbb{R}^{n_K \times C_K }$ for the next step, which can be defined as
\begin{align}
{F^{({l_K})}} = [f_1^{({l_K})},...,f_i^{({l_K})},...,f_{{n_K}}^{({l_K})}] \in {{\mathbb{R}}^{{n_K} \times {C_K}}}.
\end{align} 

In the second step, another $l_{K+1}$ SA operation can be used to perform corner aggregation for the perspective-invariant local point cloud feature generation. For details, the proposal bounding box is decoded into eight corner coordinates. To be consistent with the local point cloud coordinates, the corner coordinates, such as $(\pm l/2,\pm w/2,\pm h/2)$, are also used in the canonical reference frame, where $(l,w,h)$ represents the size of the proposal bounding box. Then, we directly set these eight corners as the $n_{K+1}$ intermediate points $X^{(l_{K+1})}$ instead of points sampled by FPS, which means that $n_{K+1}=8$. The $m_{K+1}$ neighboring points within the radius $r_{K+1}$ are also grouped and normalized for PointNet processing. After the aggregation from eight perspectives of corners, a perspective-invariant local point cloud feature $F^{(inv)} \in \mathbb{R}^{8 \times C_{K+1} }$, which can perceive the full view of an object is obtained as
\begin{align}
{F^{(inv)}} = [f_1^{({l_{K + 1}})},....,f_8^{({l_{K + 1}})}] \in {{\mathbb{R}}^{8 \times {C_{K + 1}}}}.
\end{align}

\textbf{Perspective-channel attention}. The attention mechanism is introduced to weight the contribution of each perspective differently as the third step of the ACA module, which is based on the observation that the number of points near each corner is different. Attention mechanisms have been proven to strengthen the learning of neural networks, and are widely used in machine translation \cite{attention_translation}, image caption \cite{attention_cap_tip,attention_cap_tmm}, action recognition \cite{attention_action_tip, attention_action_tmm}, image classification \cite{SENET,CBAM}, etc. As a pioneer of the 3D detector, TANet \cite{tanet_arxiv} computes the attention for point cloud voxels and achieves performance improvement with the aid of robustness. As illustrated in Fig. \ref{PCattention}, the perspective-wise and channel-wise attentions are adaptively learned from $F^{(inv)}$, inspired by \cite{tanet_arxiv}. 

For perspective-wise attention, we first exploit a max-pooling operation along the perspective-axis to obtain the most expressive perspective-wise responses $E \in \mathbb{R}^{8 \times 1}$. To explore the different contributions of perspectives, two fully connected layers with weight parameters $W_{P1} \in \mathbb{R}^{{\frac{8}{r}} \times 8}$ and $W_{P2} \in \mathbb{R}^{8 \times \frac{8}{r}}$ are utilized to learn the perspective-wise attention $S \in \mathbb{R}^{8 \times 1}$ as
\begin{align}
S = {W_{P2}}\delta ({W_{P1}}E),
\end{align}
where $\delta$ is the ReLU activation function. Similarly, channel-wise attention can also be obtained to strengthen the important channels as
\begin{align}
{T^T} = {W_{C2}}\delta ({W_{C1}}{U^T}),
\end{align}

where the most expressive channel-wise responses $U \in \mathbb{R}^{1 \times C_{K+1}}$ are aggregated by a max-pooling operation along the channel-axis. The channel-wise attention $T \in \mathbb{R}^{1 \times C_{K+1}}$ is encoded by another two fully connected layers with weight parameters $W_{C1} \in \mathbb{R}^{\frac{C_{K+1}}{r} \times C_{K+1}}$ and $W_{C2} \in \mathbb{R}^{C_{K+1} \times {\frac{C_{K+1}}{r}}}$. 
In practice, the reduction ratio $r$ is set as 1. Then, we obtain the perspective-channel attention matrix $M \in \mathbb{R}^{8 \times C_{K+1}}$, which combines the perspective-wise attention $S \in \mathbb{R}^{8 \times 1}$ and the channel-wise attention $T \in \mathbb{R}^{1 \times C_{K+1}}$ together through a matrix multiplication, as
\begin{align}
M = \sigma (S \times T),
\end{align}
where the attention values are normalized to [0, 1] by $\sigma (\cdot)$ (e.g., \textit{sigmoid} function). Thus, a reweighted feature $F^{(att)} \in \mathbb{R}^{8 \times C_{K+1}}$ can be obtained by element-wise multiplication ($\odot$) with $F^{(inv)}$ as
\begin{align}
{F^{(att)}} = M \odot {F^{(inv)}},
\end{align}
where $M$ appropriately weights the important information across the perspective-wise and channel-wise dimensions. The final output feature $F^{(ACA)} \in \mathbb{R}^{1 \times C_{K+1}}$ for each proposal from the ACA module is the sum of the eight perspectives in $F^{(att)}$:
\begin{align}
{F^{(ACA)}} = \sum\limits_{i = 1}^8 {f_i^{(att)}}.
\end{align}

\begin{figure}[!t]
	\centering
	\includegraphics[width=\linewidth]{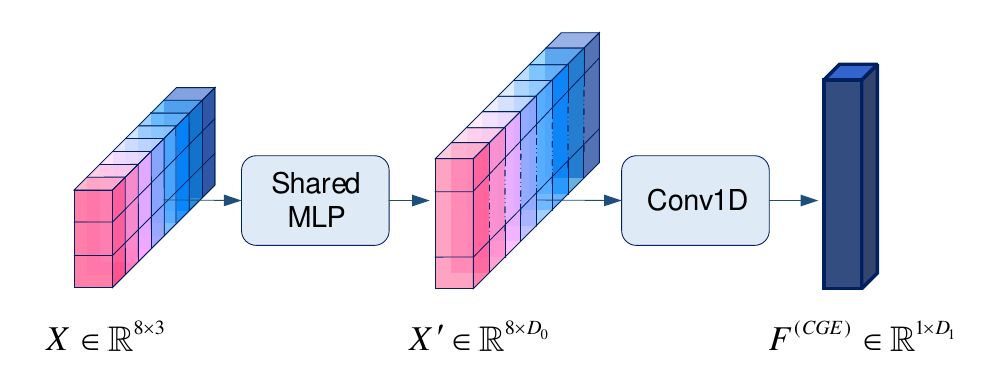}%
	\caption{Illustration of Corner Geometry Encoding module.}
	\label{CGEmodule}
\end{figure}

\subsubsection{Corner geometry encoding}
The purpose of this module is to exploit the additional geometry information provided by the proposal bounding box. Due to the preference for the detected object, most detectors only utilize the features pooled by a proposal bounding box and ignore the bounding box geometry information. However, the bounding box itself is a crucial clue for IoU prediction and free for feature extraction. As illustrated in Fig. \ref{CGEmodule}, the geometric corners feature is encoded as 
\begin{align}
{F^{(CGE)}=C({MLP}_{CGE}(X))},
\end{align}
where $MLP_{CGE} (\cdot)$ is an MLP network that embeds the corners’ absolute coordinates $X \in \mathbb{R}^{8 \times 3}$ into ${X'} \in \mathbb{R}^{8 \times D_0}$, and $C(\cdot)$ is a Conv1D with kernel $\theta \in \mathbb{R}^{D_1 \times D_0 \times 8}$ to convert $X'$ into the output $F^{(CGE)} \in \mathbb{R}^{1 \times D_1}$.

Note that we use the absolute coordinates of corners rather than the canonical coordinates, which are intended to represent the absolute position information. The experimental results (as reported in Table \ref{effect_iou_sensitive}) show that the geometry information of corners does contribute to IoU prediction. 

Besides, another MLP network $MLP_{MER}(\cdot)$ is applied to merge the concatenated features of $F^{(ACA)}$ and $F^{(CGE)}$ further, so the final IoU sensitive feature can be represented as
\begin{align}
F = ML{P_{MER}}(CAT({F^{(ACA)}},{F^{(CGE)}})).
\end{align}
As a result, both the IoU sensitive feature $F$ and the bounding box of each proposal are used for the subsequent multi-head detection prediction. 

\subsection{Detection Prediction}
\subsubsection{Box estimation branches}
In addition to the commonly used classification and regression branches, we add another branch of IoU perception. Two sets of fully connected layers are used for classification and refinement offsets regression. If its IoU is above the positive threshold $\theta_{pos}$ or below the negative threshold $\theta_{neg}$, a proposal is considered either positive or negative, respectively. Among all these positive proposals, their IoUs lack discrimination, but the location accuracy is extremely sensitive to IoU. Inspired by this, another set of fully connected layers with the same structure is added to predict IoUs. Therefore, the proposed 3D IoU-Net consists of three heads sharing the IoU sensitive features $F$ pooled by proposals.

\begin{figure}[!t]
	\centering
	\includegraphics[width=\linewidth]{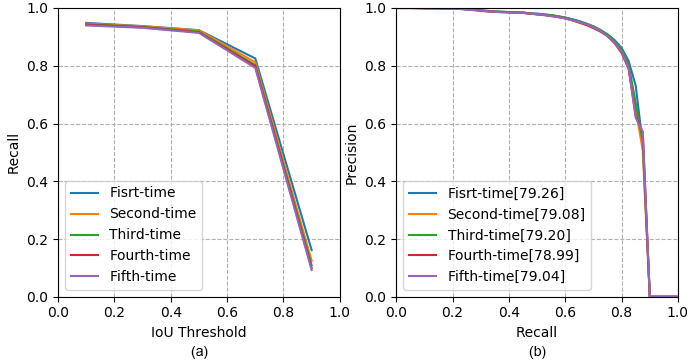}%
	\caption{Comparison of updating regression head with aligned IoU in several times. For updating in each time, we plot the recall at different IoU thresholds before NMS in (a) and the precision-recall curves on the most important moderate subset of the KITTI \textit{val} split set in (b). The average precision is shown in the square bracket in (b).}
	\label{prcurve}
\end{figure}

\subsubsection{Removing redundant results with IoU alignment}
When the NMS algorithm is applied to suppress the redundant final detection bounding boxes, the predicted IoU is expected to replace classification confidence as the ranking metric, because the localization accuracy is more sensitive to IoU. Thus, the position offsets of the proposals before and after refinement regression cannot be neglected. The IoU alignment operation aligns the IoU prediction values to the detected bounding boxes after refinement, which is inspired by the feature offsets elimination \cite{IoUUniform}. Such IoU alignment is shown as the red loop in Fig. \ref{Framework}. The final output detection results are set as the new proposals to obtain the new proposal features through our ACA and CGE modules. Then the new proposal features are sent to the IoU prediction branch for a second inference to obtain the matched IoU prediction values. Note that the regression head is only used for the new IoU prediction, and not updated in the second inference. Thus the new IoU prediction is matched to the bounding box regression through one-shot looping. For the next step, the negative result samples can be filtered out by classification. Then, the redundant bounding boxes among the positive result samples are removed by IoU-guided NMS in the post-processing procedure. Due to the IoU alignment at the inference stage, we further improve the performance of IoU prediction and the average precision of 3D detection without any extra training.

\textbf{Why not update the bounding box regression}? Since the updated IoU prediction is computed from the output of the old bounding box regression head, the updated IoU prediction is aligned with the old bounding box regression. If the bounding box regression is also updated, there still exists mismatching between the new bounding box regression and the new IoU prediction. Besides, as shown in Fig. \ref{prcurve}(a), when updating the bounding box regression head and IoU prediction head in several loops, the recall value at each loop is not improved. This is because the refinement of regression has a limit. As a result, the average precision shown in Fig. \ref{prcurve}(b) is not continuously improved. Therefore, we resolve the mismatching by only updating the IoU prediction through one-shot looping.

\subsection{Loss Function}
The loss is composed of a classification loss, refine regression loss, IoU loss and auxiliary corner loss as 
\begin{align}
{L_{total}} = \alpha {L_{cls}} + \beta {L_{reg}} + \gamma {L_{iou}} + \lambda {L_{aux}},
\end{align}
where $\alpha$, $\beta$, $\gamma$, and $\lambda$ are constants (e.g., 1, 1, 20, 1) that weight each loss term of multi-task learning. $L_{cls}$ and $L_{reg}$ are defined the same as in \cite{PointRCNN}. The IoU loss is employed to optimize the IoU prediction branch as
\begin{align}
{L_{iou}} = \frac{1}{N}\sum\limits_i {[Io{U_i} \geqslant 0.55]{L_{Smooth - L1}}({A_i},{G_i})},
\end{align}
where the Iverson bracket indicator function $[{IoU}_{i} \geqslant 0.55]$ reaches 1 when $IoU_i \geqslant 0.55$ and $0$ otherwise, $N$ is the number of proposals that satisfy the condition $IoU_i \geqslant 0.55$, $G_i$ is the calculated IoU between the $i$-$th$ proposal and the corresponding ground truth bounding box, and $A_i$ is the predicted IoU. The auxiliary loss $L_{aux}$ is the $L_2$ distance between the predicted eight corners and the matched ground truth, which can be expressed as
\begin{align}
{L_{aux}} = \frac{1}{N}\sum\limits_i {([Io{U_i} \geqslant 0.55]\sum\limits_{k = 1}^8 {\left\| {{P_{ik}} - {G_{ik}}} \right\|)} }
\end{align}
where $P_{ik}$ and $G_{ik}$ are the predicted and the ground truth location of the $i$-$th$ proposal’s corner $k$.

\section{Experiments}
In this section, we first describe the experimental setup and then compare the proposed 3D IoU-Net model with state-of-the-art methods on the widely used KITTI dataset \cite{KITTIDataset}. Several extensive ablation studies are conducted to validate the effectiveness of each component in the proposed 3D IoU-Net.

\subsection{Experimental Setup}
\subsubsection{Dataset}
The KITTI dataset \cite{KITTIDataset} provides 7481 training images/point clouds and 7518 testing images/point clouds for 3D object detection of cars, cyclists and pedestrians. For each category, three difficulty levels are involved (Easy, Moderate and Hard), which depend on the size, occlusion level and truncation of 3D objects. The proposed model is evaluated on the class Car, due to its large amount of data and complex scenarios. Moreover, most of state-of-the-art methods only test their models on this class. Detection performance is generally compared with the official KITTI evaluation metrics, which are 3D detection average precision (3D AP) and bird’s eye view detection average precision (BEV AP). The training samples are provided with labels, while the results on the \textit{test} set must be submitted to the official test server \cite{KITTIleaderboard} for evaluation. Since the ground truth for the test samples is not available, training samples are generally divided into the \textit{train} split set (3712) and the \textit{val} split set (3769) for training and validation, respectively.
\subsubsection{Training details}
The RPN is leveraged from PointRCNN \cite{PointRCNN}. The ADAM optimizer with an initial learning rate of 0.001 is applied for training the remaining parameters from scratch. We train over 30 epochs on one GTX1080Ti GPU with batch size 4. Each batch consists of 64 proposals, which are sampled from one input point cloud with ratio 1:1 for positive and negative proposals. For the proposals used for classification branch training, the positive threshold $\theta_{pos}$ and the negative threshold $\theta_{neg}$ are set to 0.6 and 0.45, respectively. Only proposals with an IoU larger than 0.55 are considered for training the regression and IoU prediction branches. To avoid overfitting, in addition to typical data augmentation like rotating around the \textit{Z}-axis, flipping along the \textit{X}-axis, and global scaling in the LiDAR coordinate system, we also augment the dataset using some ground truth bounding boxes copied from other frames, like in \cite{Second}. Then, we randomly select 16k points from the entire point cloud for each scene to align the network input. We randomly sample or repeat so that the number of points inside a proposal is 512.

When evaluating the performance on the \textit{test} set with the KITTI official test server, the model is trained on our own random split train/val set at a ratio of 4:1. Apart from this, all models used in the performance comparisons and ablation studies are trained on the common \textit{train} split set and evaluated on the common \textit{val} split set.


\begin{table*}[!t]
	\caption{Performance comparison on Car class of the KITTI \textit{test} set by submitting to the official test server.}
	\centering
	\label{APcomparison_test}
	\begin{threeparttable}[b]
		
	\begin{tabular}{ccccccccc}
		\toprule
		\multirow{2}{*}{Method} & \multirow{2}{*}{Reference} & \multirow{2}{*}{Modality} & \multicolumn{3}{c}{3D Object Detection AP}       & \multicolumn{3}{c}{BEV Object Detection AP}      \\
		&                            &                           & Easy           & Moderate       & Hard           & Easy           & Moderate       & Hard           \\
		\midrule
		MV3D \cite{MV3D}           & CVPR 2017                  & RGB + LiDAR               & 74.97          & 63.63          & 54.00          & 86.62          & 78.93          & 69.80          \\
		AVOD-FPN \cite{AVOD}       & IROS 2018                  & RGB + LiDAR               & 83.07          & 71.76          & 65.73          & 90.99          & 84.82          & 79.62          \\
		Conti-Fuse \cite{Contifuse}     & ECCV 2018                  & RGB + LiDAR               & 83.68          & 68.78          & 61.67          & 94.07          & 85.35          & 75.88          \\
		UberATG-MMF \cite{MMF}    & CVPR 2018                  & RGB + LiDAR               & \textbf{88.40} & 77.43          & 70.22          & 93.67          & 88.21          & 81.99          \\
		PI-RCNN \cite{PIRCNN}        & AAAI 2020\tnote{1}                  & RGB + LiDAR               & 84.37          & 74.82          & 70.03          & 91.44          & 85.81          & 81.00          \\
		SECOND \cite{Second}         & Sensors 2018               & LiDAR                     & 83.34          & 72.55          & 65.82          & 89.39          & 83.77          & 78.59          \\
		PointPillars \cite{PointPillars}   & CVPR 2019                  & LiDAR                     & 82.58          & 74.31          & 68.99          & 90.07          & 86.56          & 82.81          \\
		3D IoU Loss \cite{3DIoUloss}    & 3DV 2019                   & LiDAR                     & 86.16          & 76.50          & 71.39          & 91.36          & 86.22          & 81.20          \\
		PointRCNN \cite{PointRCNN}      & CVPR 2019                  & LiDAR                     & 86.96          & 75.64          & 70.70          & 92.13          & 87.39          & 82.72          \\
		Fast PointRCNN \cite{Fast-PointRCNN} & ICCV 2019                  & LiDAR                     & 85.29          & 77.40          & 70.24          & 90.87          & 87.84          & 80.52          \\
		Part-A2 \cite{PartA2_TPAMI}        & TPAMI 2020\tnote{2}                 & LiDAR                     & 87.81          & 78.49          & \textbf{73.51} & 91.70          & 87.79          & \textbf{84.61} \\
		STD \cite{STD}            & ICCV 2019                  & LiDAR                     & 87.95          & \textbf{79.71} & \textbf{75.09} & \textbf{94.74} & \textbf{89.19} & \textbf{86.42} \\
		PointRGCN \cite{PointRGCN}      & Arxiv 2020                 & LiDAR                     & 85.97          & 75.73          & 70.60          & 91.63          & 87.49          & 80.73          \\
		TANet \cite{tanet_arxiv}          & AAAI 2020\tnote{1}                  & LiDAR                     & 84.39          & 75.94          & 68.82          & 91.58          & 86.54          & 81.19          \\
		3D IoU-Net (Ours)       & -                          & LiDAR                     & \textbf{87.96} & \textbf{79.03} & 72.78          & \textbf{94.76} & \textbf{88.38} & 81.93 \\        
		\bottomrule
	\end{tabular}

	\begin{tablenotes}
	\item[1] Accepted by AAAI 2020
	\item[2] Accepted by TPAMI 2020
	\end{tablenotes}

	\end{threeparttable}	
\end{table*}


\begin{table*}[!t]
	\caption{Performance comparison on Car class of the KITTI \textit{val} split set.}
	\centering
	\label{APcomparison_val}
		
	\begin{tabular}{ccccccc}
		\toprule
		\multirow{2}{*}{Method} & \multirow{2}{*}{Reference} & \multirow{2}{*}{Modality} & \multicolumn{3}{c}{3D Object Detection AP}       &  \multirow{2}{*}{mAP}               \\
		&                            &                           & Easy           & Moderate       & Hard           &             \\
		\midrule
		MV3D\cite{MV3D}            & CVPR 2017                  & RGB + LiDAR               & 71.29          & 62.68          & 56.56          & 63.51          \\
		AVOD-FPN \cite{AVOD}       & IROS 2018                  & RGB + LiDAR               & 84.41          & 74.44          & 68.65          & 75.83          \\
		ContFuse \cite{Contifuse}       & ECCV 2018                  & RGB + LiDAR               & 86.32          & 73.25          & 67.81          & 75.79          \\
		UberATG-MMF \cite{MMF}    & CVPR 2018                  & RGB + LiDAR               & 87.90          & 77.86          & 75.57          & 80.44          \\
		PI-RCNN \cite{PIRCNN}        & AAAI 2020                  & RGB + LiDAR               & 87.63          & 77.87          & 76.17          & 80.56          \\
		SECOND \cite{Second}         & Sensors 2018               & LiDAR                     & 87.43          & 76.48          & 69.10          & 77.67          \\
		3D IoU Loss \cite{3DIoUloss}    & 3DV 2019                   & LiDAR                     & 89.16          & 78.33          & 77.25          & 81.58          \\
		PointRCNN \cite{PointRCNN}      & CVPR 2019                  & LiDAR                     & 88.88          & 78.63          & 77.38          & 81.63          \\
		Fast PointRCNN \cite{Fast-PointRCNN} & ICCV 2019                  & LiDAR                     & 89.12          & 79.00          & 77.48          & 81.87          \\
		Part-A2 \cite{PartA2_TPAMI}        & TPAMI 2020                 & LiDAR                     & \textbf{89.47} & \textbf{79.47} & 78.54          & \textbf{82.49} \\
		STD \cite{STD}            & ICCV 2019                  & LiDAR                     & \textbf{89.70} & \textbf{79.80} & \textbf{79.30} & \textbf{82.93} \\
		PointRGCN \cite{PointRGCN}      & Arxiv 2020                 & LiDAR                     & 88.37          & 78.54          & 77.60          & 81.50          \\
		TANet \cite{tanet_arxiv}          & AAAI 2020\tnote{1}                  & LiDAR                     & 87.52          & 76.64          & 73.86          & 79.34          \\
		3D IoU-Net (Ours)       & -                          & LiDAR                     & 89.31          & 79.26          & \textbf{78.68} & 82.42         \\
		\bottomrule
	\end{tabular}



\end{table*}

\subsection{Results on KITTI Dataset}
We compare the proposed 3D IoU-Net with the state-of-the-art 3D detectors on both the \textit{test} set and the \textit{val} split set of the KITTI dataset. Table \ref{APcomparison_test} shows the performance on the \textit{test} set from the official online leaderboard as of Mar. 1th 2020, where top-2 is marked in bold. For the most competitive metric, 3D AP, 3D IoU-Net achieves stronger performance than most previous 3D detectors and outperforms the best multi-sensor detector UberATG-MMF \cite{MMF} by 1.60\% on the moderate level and 2.56\% on the hard level. Note that both PI-RCNN \cite{PIRCNN} and PointRGCN \cite{PointRGCN} are based on PointRCNN, but neither of them reproduces its performance, while our 3D IoU-Net completely surpasses all of them. The proposed 3D IoU-Net is slightly better than STD \cite{STD} on the easy level AP of 3D and BEV detection. For the moderate level AP both on 3D and BEV detection, the absolute value of the gap between our method and STD is also less than 1\%. 

Table \ref{APcomparison_val} shows the 3D AP comparison on the \textit{val} split set for reference, using the data provided in each method’s paper. The mAP represents the mean average precision of each difficulty level. Our 3D IoU-Net is still competitive, demonstrating that our model has better generalization ability on the \textit{test} set. 

\subsection{Ablation Studies}
In this section, several extensive ablation experiments are conducted to analyze the effects of individual components of 3D IoU-Net and discuss our design choices. All ablation experiment results are evaluated on the \textit{val} split set and follow the official evaluation protocol. All models used are trained on the \textit{train} split with the same training schedule.

\subsubsection{Analysis of alignment operation}
\begin{figure*}[!t]
	\centering
	\includegraphics[width=\linewidth]{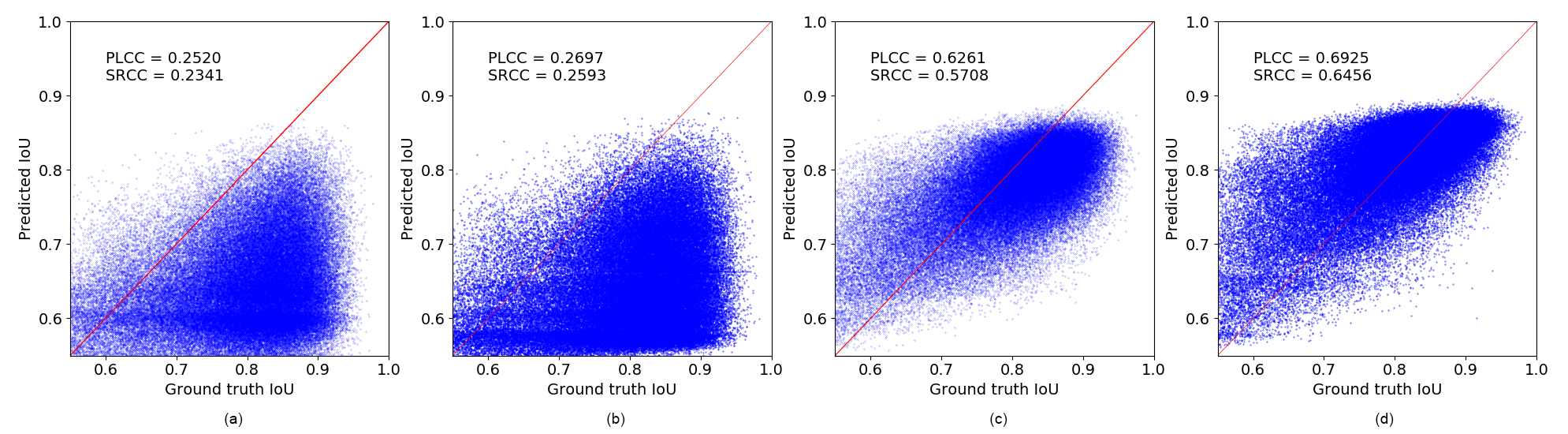}%
	\caption{The correlation between the predicted IoU and the corresponding ground truth IoU. (a) Baseline model without IoU alignment. (b) 3D IoU-Net without IoU alignment. (c) Baseline model with IoU alignment. (d) 3D IoU-Net with IoU alignment.}
	\label{correlation}
\end{figure*}

To explore the improvement that the IoU alignment brings to the 3D detection mAP, we conduct experiments on two models. One is PointRCNN with an added IoU prediction head as our baseline model, and the other is our 3D IoU-Net. Table \ref{effect_iou_align} shows the performance improvements of the two models, both of which use the predicted IoU as the detection confidence of the NMS post-processing and the AP evaluation. As can be seen, the 3D mAP improvements are 3.36\% and 4.58\%, respectively, which implies that the models benefit from the elimination of assignment mismatching by IoU alignment. Fig. \ref{correlation} provides a visualization of the correlation between the IoU prediction value and the ground truth value, considering the detected bounding boxes that have an IoU above 0.55 with the corresponding ground truth bounding box before NMS, on the val split set. Pearson’s Linear Correlation Coefficient (PLCC) and Spearman’s Rank-order Correlation Coefficient (SRCC) describe the strength of IoU prediction from the perspective of linear correlation and order correlation, respectively. We can see that the IoU alignment contributes to predicting a more relevant detection confidence to the ground-truth IoU.

In Table \ref{effect_class_align}, whether or not alignment improves the classification confidence is further explored. Unfortunately, for both two models, the 3D mAP with the aligned classification confidence as the detection confidence is worse than that using the original classification confidence, which means that the alignment operation is not suitable for classification confidence. A proposal is considered positive during training if its IoU with the matched ground-truth bounding box is larger than a given threshold. This lacks discriminative learning among all positive samples. To make matters worse, most of the refined bounding boxes are positive samples. Such an imbalance has a negative impact on classification confidence alignment.

For the baseline model with only the IoU alignment, its 3D mAP still lags behind the classification confidence-based 3D mAP by 0.44\%, as shown in the second row in Table \ref{effect_iou_align} and the first row in Table \ref{effect_class_align}. The reason lies in the lack of IoU sensitive feature learning. When it comes to our 3D IoU-Net, the aligned IoU-based mAP overtakes the classification confidence based 3D mAP of the baseline model by 0.99\%, as shown in the last row in Table \ref{effect_iou_align} and the first row in Table \ref{effect_class_align}. This performance gap suggests that the IoU alignment complements the IoU sensitive feature learning in our 3D IoU-Net. 

\begin{table}[!t]
	\caption{Abation study on the effects of IoU alignment.}
	\centering
	\label{effect_iou_align}
	\begin{tabular}{cccccc}
		\toprule	
		\multirow{2}{*}{Model} & \multirow{2}{*}{Alignment} & \multicolumn{3}{c}{3D Object Detection AP}       & \multirow{2}{*}{mAP} \\
		&                            & Easy           & Mod.           & Hard           &                      \\
		\midrule
		Baseline               & $\times$                          & 83.85          & 74.94          & 74.10          & 77.63                \\
		Baseline               & $\surd$                          & \textbf{87.60} & \textbf{78.16} & \textbf{77.21} & \textbf{80.99}       \\
		3D IoU-Net             & $\times$                          & 83.78          & 75.08          & 74.66          & 77.84                \\
		3D IoU-Net             & $\surd$                          & \textbf{89.31} & \textbf{79.26} & \textbf{78.68} & \textbf{82.42}       \\
		\bottomrule
	\end{tabular}
\end{table}

\begin{table}[!t]
	\caption{Abation study on the effects of classification confidence alignment.}
	\centering
	\label{effect_class_align}
	\begin{tabular}{cccccc}
		\toprule
		\multirow{2}{*}{Model} & \multirow{2}{*}{Alignment} & \multicolumn{3}{c}{3D Object Detection AP}       & \multirow{2}{*}{mAP} \\
		&                            & Easy           & Mod.           & Hard           &                      \\
		\midrule
		Baseline               & $\times$                          & \textbf{88.12} & \textbf{78.52} & \textbf{77.66} & \textbf{81.43}       \\
		Baseline               & $\surd$                          & 86.95          & 77.47          & 76.40          & 80.27                \\
		3D IoU-Net             & $\times$                          & \textbf{88.51} & \textbf{78.67} & \textbf{77.90} & \textbf{81.69}       \\
		3D IoU-Net             & $\surd$                          & 87.38          & 77.73          & 76.99          & 80.70               \\
		\bottomrule
	\end{tabular}
\end{table}

\subsubsection{Effects of IoU sensitive feature learning}
Table \ref{effect_iou_sensitive} demonstrates a further investigation on the importance of the ACA module and the CGE module for IoU sensitive feature learning. All the models use aligned IoU as the detection confidence for fair comparison, and the first row in Table \ref{effect_iou_sensitive} is set as the new baseline. With only the ACA module and only the CGE module, the 3D mAP is boosted to 82.15\% and 81.97\%, respectively. As shown in the last row in Table \ref{effect_iou_sensitive}, when combining them together, our model yields the best 3D mAP of 82.42\%, outperforming the baseline by 1.43\%. It should be noted that both of the proposed two modules bring improvements over the baseline. As can be observed, combining the two modules achieves the best PLCC and SRCC, suggesting that the perspective-invariant local point cloud feature aggregated by the ACA module and the bounding box geometry feature encoded by the CGE module both contribute to IoU sensitive feature learning.

In addition, even without the alignment of the predicted IoU or the classification confidence, the 3D IoU-Net still outperforms the baseline model. This can be concluded from the comparison between the first row and the third row in Table \ref{effect_iou_align} and the comparison between the first row and the third row in Table \ref{effect_class_align}. The IoU sensitive feature obtained from the ACA and CGE modules is beneficial not only to the IoU prediction but also to the classification.

\subsubsection{Analysis of ACA module}
To verify the effectiveness of the corner aggregation and attentive feature extraction in the ACA module, we provide three alternative models, which aggregate the local point cloud feature in different ways. Table \ref{effect_aca_module} shows that the performance drops when replacing eight corners with eight random points sampled by FPS for aggregation, which validates that the proposed corner aggregation can learn a more perspective-invariant feature representation to capture the full view of an object. Moreover, in the first and the second row of Table \ref{effect_aca_module}, when compared to the models that treat each corner and channel equally, the performance increases slightly when the extracted features are reweighted by the perspective-wise and channel-wise attention. The different distribution of points makes them contribute differently.

\begin{table}[!t]
	\caption{Ablation study on the effects of IoU sensitive feature generation.}
	\centering
	\label{effect_iou_sensitive}
	\begin{tabular}{cccccccc}
		\toprule
		\multirow{2}{*}{ACA} & \multirow{2}{*}{CGE} & \multicolumn{3}{c}{3D Object Detection AP}       & \multirow{2}{*}{mAP} & \multirow{2}{*}{PLCC} & \multirow{2}{*}{SRCC} \\
		&                      & Easy           & Mod.           & Hard           &                      &                       &                       \\
		\midrule
		$\times$                    & $\times$                    & 87.60          & 78.16          & 77.21          & 80.99                & 0.6261                & 0.5708                \\
		$\surd$                    & $\times$                    & 89.29          & 78.98          & 78.18          & 82.15                & 0.6795                & 0.6337                \\
		$\times$                    & $\surd$                    & 88.70          & 78.80          & 78.41          & 81.97                & 0.6773                & 0.6325                \\
		$\surd$                    & $\surd$                    & \textbf{89.31} & \textbf{79.26} & \textbf{78.68} & \textbf{82.42}       & \textbf{0.6925}       & \textbf{0.6456}      \\
		\bottomrule
	\end{tabular}
\end{table}

\begin{table}[!t]
	\caption{Ablation study on the individual components of ACA module.}
	\centering
	\label{effect_aca_module}
	\begin{tabular}{cccccc}
		\toprule
		\multirow{2}{*}{Aggregation} & \multirow{2}{*}{Merge} & \multicolumn{3}{c}{3D Object Detection AP}       & \multirow{2}{*}{mAP} \\
		&                        & Easy           & Mod.           & Hard           &                      \\
		\midrule
		Corners                      & Attentive              & \textbf{89.31} & \textbf{79.26} & \textbf{78.68} & \textbf{82.42}       \\
		Random                       & Attentive              & 89.06          & 78.98          & 78.46          & 82.17                \\
		Corners                      & Mean                   & 89.27          & 79.07          & 78.41          & 82.25                \\
		Random                       & Mean                   & 88.91          & 78.80          & 78.33          & 82.01               \\
		\bottomrule
	\end{tabular}
\end{table}

\begin{figure*}[t]
	\centering
	\includegraphics[width=\linewidth]{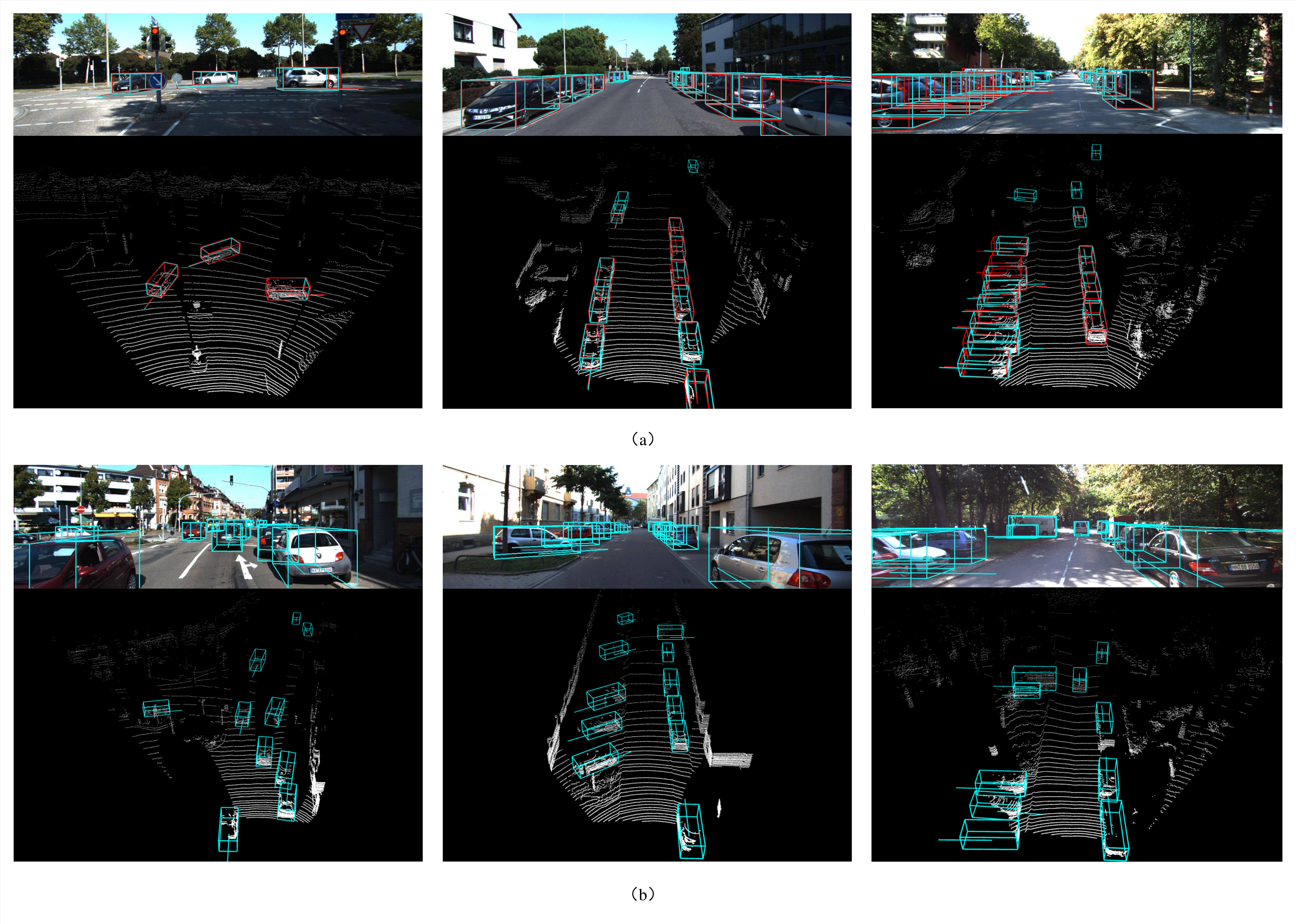}%
	\caption{Qualitative detection results of 3D IoU-Net. (a) The results on the \textit{val} split set. (b) The results on the \textit{test} set. Detected bounding boxes and corresponding ground truth are colored cyan and red respectively. The lines extending from the bottom of bounding boxes point to objects’ heading direction.}
	\label{Quantitive}
\end{figure*}

\subsection{Qualitative Results}
Several qualitative detection results of the proposed 3D IoU-Net on both the \textit{val} split set and the \textit{test} set of KITTI dataset are presented in Fig. \ref{Quantitive} for visualization. Note that the RGB image is only used for better viewing. The 3D IoU-Net only takes the point cloud as input and outputs the 3D bounding boxes as the detection results. With the ground truth provided on the \textit{val} split set, it can be observed that the detection results for object instances can well overlap with the ground truth bounding boxes in any orientation. Our model can perform well for complex scenes with both left-right and front-back heading vehicles, as shown in the third column in Fig. \ref{Quantitive}. For the \textit{test} set, viewing with the camera image and the LiDAR point cloud, we can also observe that it still accurately locates the objects.

\section{Conclusion}
In this paper, we proposed a novel 3D object detection framework, named 3D IoU-Net, for accurate 3D object localization with contributions in IoU sensitive feature learning and resolving IoU assignment mismatching. The Attentive Corners Aggregation (ACA) module and the Corner Geometry Encoding (CGE) module are crucial to IoU sensitive feature learning. The former attentively aggregates a local point cloud feature from the perspectives of eight bounding box corners. The latter provides extra corner geometry information for enhanced IoU sensitive feature learning. Besides, the IoU alignment operation is key to eliminating IoU assignment mismatching and enhancing the performance of 3D object detection. With more accurate detection confidence from IoU perception, we reduce the incorrect removals during post-processing. The comprehensive evaluation shows that our method achieves state-of-the-art performance.

\bibliographystyle{IEEEtran}
\bibliography{ljl_3DIoUNet}
	
\end{document}